%% file: PaperForReview.tex
\documentclass[10pt,twocolumn,letterpaper]{article}

\usepackage[pagenumbers]{cvpr} 

\usepackage{graphicx}
\usepackage{amsmath}
\usepackage{amssymb}
\usepackage{booktabs} \def\ul{}
\usepackage{times}
\usepackage{epsfig}
\usepackage{mathrsfs}
\usepackage{bm}
\usepackage{float}
\usepackage{multirow}
\usepackage{enumitem}
\usepackage{algorithm}
\usepackage{algorithmic}
\usepackage{arydshln}
\usepackage[figuresright]{rotating}
\usepackage[table,xcdraw]{xcolor}
\usepackage[accsupp]{axessibility}  

\usepackage[normalem]{ulem}
\useunder{\uline}{\ul}{}

\DeclareMathOperator*{\argtopk}{arg-topk}

\usepackage[pagebackref,breaklinks,colorlinks]{hyperref}

\usepackage[capitalize]{cleveref}
\crefname{section}{Sec.}{Secs.}
\Crefname{section}{Section}{Sections}
\Crefname{table}{Table}{Tables}
\crefname{table}{Tab.}{Tabs.}

\def\cvprPaperID{6649} 
\def\confName{CVPR}
\def\confYear{2023}

\begin{document}

\title{Robust Multiview Point Cloud Registration with Reliable \\ Pose Graph Initialization and History Reweighting}

\author{
Haiping Wang\thanks{Both authors contribute equally to this research.}\\
Wuhan University\\
\and
Yuan Liu\textsuperscript{*}\\
The University of Hong Kong\\
\and 
Zhen Dong\thanks{Corresponding authors: [dongzhenwhu, bshyang]@whu.edu.cn}\\
Wuhan University\\
\and 
Yulan Guo\\
Sun Yat-sen University\\
\and 
Yu-Shen Liu\\
Tsinghua University\\
\and 
Wenping Wang\\
Texas A\&M University\\
\and 
Bisheng Yang\textsuperscript{†}\\
Wuhan University\\
}
\maketitle

\input{content/00_abstract.tex}


\input{content/01_introduction.tex}
\input{content/02_relatedwork.tex}

\input{content/03_method.tex}
\input{content/04_experiments.tex}

\input{content/05_conclusion.tex}

\section{Acknowledgement}
This research is jointly sponsored by the National Key Research and Development Program of China (No.2022YFB3904102), the National Natural Science Foundation of China Projects (No.42171431, U20A20185, 61972435), the Open Fund of Hubei Luojia Laboratory (No.2201000054) and the Guangdong Basic and Applied Basic Research Foundation (2022B1515020103).

{\small
\bibliographystyle{ieee_fullname}
\bibliography{egbib}
}

\input{content/06_supplementary.tex}

\end{document}

%% file: content/00_abstract.tex
\begin{abstract}
In this paper, we present a new method for the multiview registration of point cloud.
Previous multiview registration methods rely on exhaustive pairwise registration to construct a densely-connected pose graph and apply Iteratively Reweighted Least Square (IRLS) on the pose graph to compute the scan poses. 
However, constructing a densely-connected graph is time-consuming and contains lots of outlier edges, which makes the subsequent IRLS struggle to find correct poses.
To address the above problems, we first propose to use a neural network to estimate the overlap between scan pairs, which enables us to construct a sparse but reliable pose graph. 
Then, we design a novel history reweighting function in the IRLS scheme, which has strong robustness to outlier edges on the graph.
In comparison with existing multiview registration methods, our method achieves $11\%$ higher registration recall on the 3DMatch dataset and $\sim13\%$ lower registration errors on the ScanNet dataset while reducing $\sim70\%$ required pairwise registrations. Comprehensive ablation studies are conducted to demonstrate the effectiveness of our designs.
The source code is available at \url{https://github.com/WHU-USI3DV/SGHR}.
\end{abstract}

%% file: content/01_introduction.tex

\section{Introduction}
\label{sec:intro}
Point cloud registration is a prerequisite for many tasks such as 3D reconstruction~\cite{guo2020deep,dong2020registration,huang2021comprehensive} and 3D segmentation~\cite{landrieu2018large,hu2020randla}. Most recent registration methods~\cite{gojcic2019perfect,bai2020d3feat,ao2021spinnet,wang2022you,huang2021predator,yu2021cofinet,li2022lepard,qin2022geometric} mainly focus on pairwise registration of two partial point clouds (scans), which can only reconstruct a part of the scene.
In order to get a completed scene reconstruction, all partial point clouds should be simultaneously aligned, which is called \textit{multiview registration}. 
Due to its complexity, multiview point cloud registration receives less attention recently and only few recent studies propose multiview registration methods~\cite{yang2016automatic,dong2018hierarchical,huang2019learning,gojcic2020learning,yew2021learning}. 

\begin{figure}
\begin{center}
\includegraphics[width=1\linewidth]{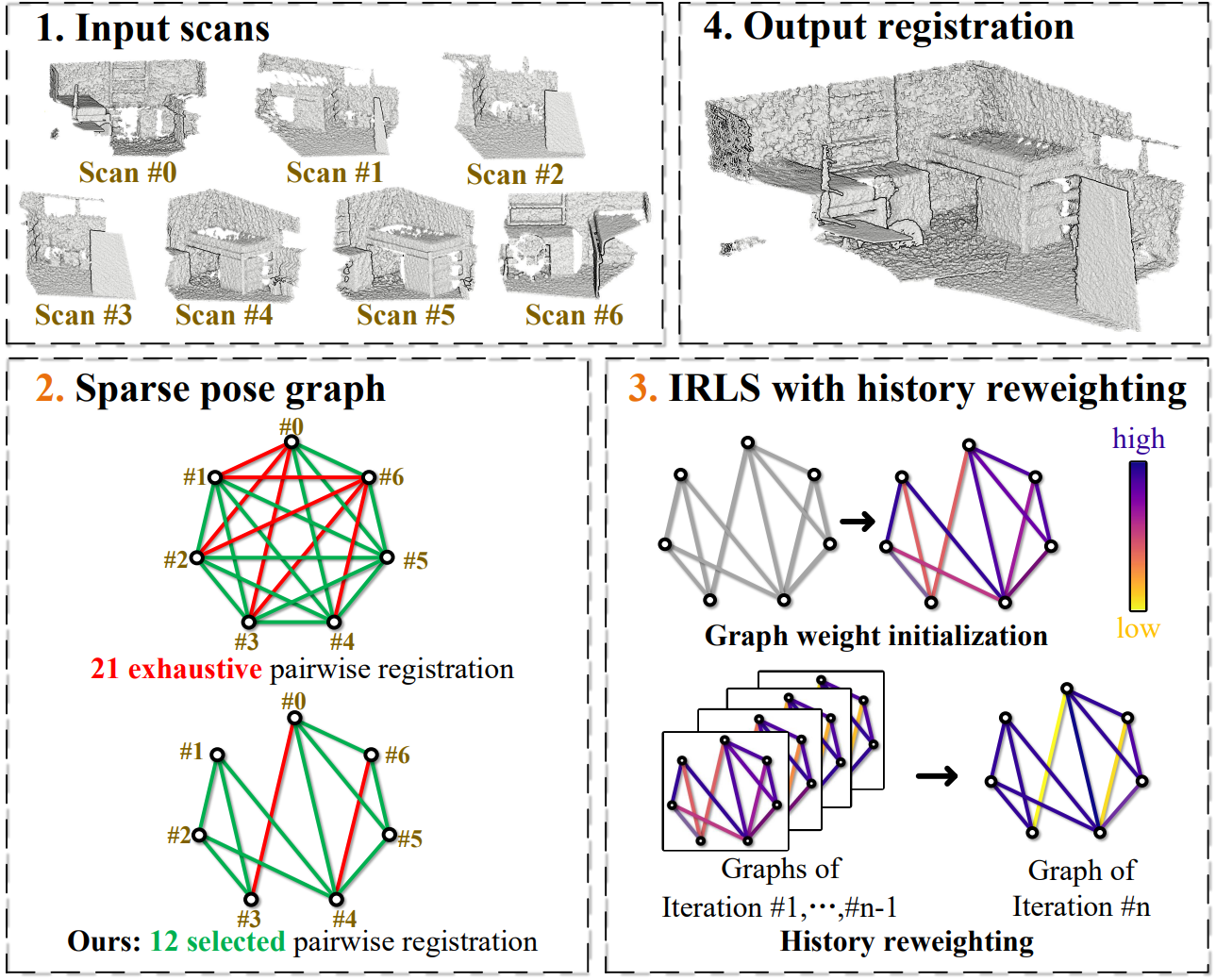}
\end{center}
\vspace{-10pt}
\caption{ \textbf{Overview}. (1) Given $N$ unaligned partial scans, our target is to register all these scans into (4) a completed point cloud. 
Our method has two contributions. (2) We learn a global feature vector to initialize a sparse pose graph which contains much less outliers and reduces the required number of pairwise registrations.
(3) We propose a novel IRLS scheme. In our IRLS scheme, we initialize weights from both global features and pairwise registrations. Then, we design a history reweighting function to iteratively refine poses, which improves the robustness to outliers.}
\label{fig:teaser}
\vspace{-15pt}
\end{figure}


Given $N$ unaligned partial point clouds, multiview registration aims to find a globally-consistent pose for every partial point cloud.
A commonly-adopted pipeline of multiview registration consists of two phases~\cite{yew2021learning}. 
First, a pairwise registration algorithm~\cite{huang2021predator,wang2022you,qin2022geometric} is applied to exhaustively estimate the relative poses of all $\binom{N}{2}$ scan pairs, which forms a fully-connected pose graph. The edges of the graph stand for the relative poses of scan pairs while nodes represent scans.
Since the dense pose graph may include inaccurate or even incorrect relative poses (outliers) between two irrelevant scans, in the second phase, these pairwise poses are jointly optimized by enforcing the cycle consistency~\cite{huang2019learning} to reject outlier edges and improve accuracy. For the second phase, most recent methods, including handcrafted methods~\cite{choi2015robust,arrigoni2016spectral,huang2017translation} or learning-based~\cite{huang2019learning,gojcic2020learning,yew2021learning} methods, follow a scheme of Iterative Reweighting Least Square (IRLS). In the IRLS, initial weights are assigned to edges to indicate these edges are reliable or not. Then, based on the weights, a synchronization algorithm is applied to compute a new relative pose on every edge. After that, the weights on edges are updated according to the difference between the old relative poses and the new ones. IRLS iteratively synchronize poses from edge weights and update weights with synchronized poses. 

In an ideal case, an IRLS scheme will gradually lower the weights of the outlier edges and only consider the inlier edges for pose synchronization. However, the initial densely-connected graph contains lots of outliers, which often prevents the iterative reweighting mechanism of IRLS from finding correct edges. To improve the robustness to outliers, many researches focus on applying advanced handcrafted reweighting functions~\cite{huang2017translation,chatterjee2017robust} or designing graph network to learn reweighting functions~\cite{huang2019learning,yew2021learning}. However, the handcrafted reweighting functions usually require a good initialization to converge to the correct poses while learning-based reweighting methods may not generalize to unseen settings. Designing a robust IRLS algorithm still remains an open problem.



In this paper, we show that multiview registration can be improved from two aspects, as shown in Fig.~\ref{fig:teaser}. First, we learn a good initialization of the input pose graph which avoids exhaustive pairwise registrations and reduces the outlier ratio. Second, we propose a novel history reweighting function which enables a stable convergence to correct poses in the IRLS scheme.

In the pose graph construction, we learn a global feature on each point cloud and the correlation of two global feature indicates the overlap ratio between two point clouds. Such global features enable us to generate a sparse pose graph with fewer but more reliable edges instead of a densely-connected graph. After that, we only need to apply the pairwise registration algorithm and IRLS on these sparse edges, which greatly reduce the computation complexity of pairwise registration from $O(N^2)$ to $O(N)$. Meanwhile, these reliable edges contain much less outliers than the fully-connected graph, which provides the possibility to find more accurate and consistent global poses in IRLS. 

\begin{figure}
\begin{center}
\includegraphics[width=\linewidth]{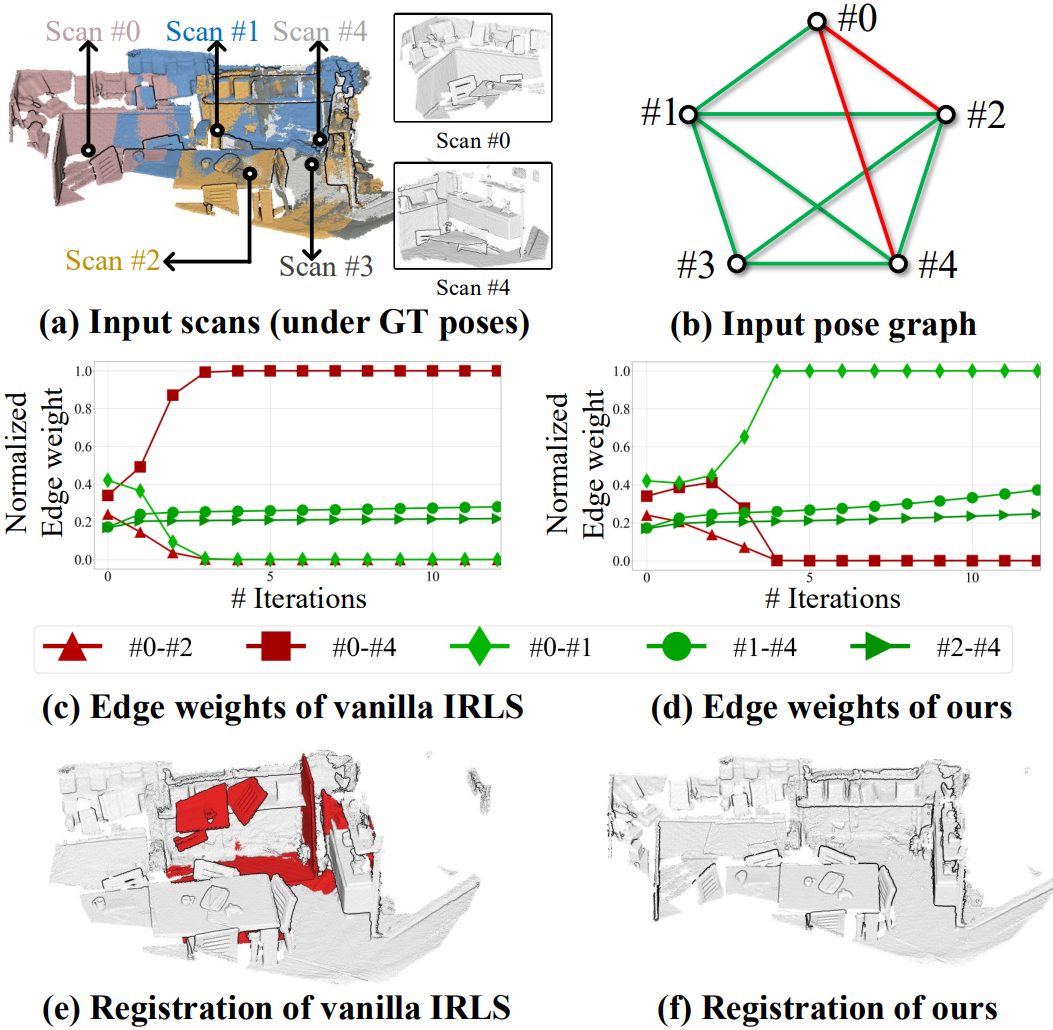}
\end{center}
\caption{An example on the 3DMatch dataset. (a) The input scans under the ground truth poses. 
(b) The constructed sparse pose graph with two incorrect relative poses (\#0-\#2 and \#0-\#4), where \#0 and \#4 looks very similar to each other so that the pose graph incorrectly include this scan pair. (c) and (d) show the normalized weights on the graph edges on different iterations of the vanilla IRLS and our method respectively. Our method is able to find the outlier edges and gradually reduce their weights while vanilla IRLS is biased towards the outlier edge (\#0-\#4) after few iterations. (e) and (f) are the multiview registration results of the vanilla IRLS and our method respectively.}
\label{fig:toy}
\vspace{-5pt}
\end{figure}

Though the initial graph contains much less outliers, existing IRLS algorithms are still sensitive to these outliers and can be totally biased towards these outliers in the first few iterations. An example is shown in Fig.~\ref{fig:toy}: the initial graph only contains two outlier edges. However, the outlier scan pair ``\#0-\#4" looks very similar and thus is initialized with a large weight. Such an incorrect large weight interferes the subsequent pose synchronization and brings systematic errors to the synchronized poses.
The vanilla IRLS trusts all synchronized poses and is easily dominated by these erroneous poses, which leads to incorrect convergence as shown in Fig.~\ref{fig:teaser}(c).
To address this problem, we propose a simple yet effective reweighting function called the \textit{history reweighting function}. In history reweighting function, edge weights at a specific iteration not only depends on the synchronized poses at the current iterations but also considers the historical synchronized poses in previous iterations, which acts like a regularizer to prevent the IRLS from being dominated by outliers at the early unstable iterations as shown in the Fig.~\ref{fig:toy}~(d). 
Then, the edge weights in our graph gradually stabilize in the subsequent iterative refinements, leading to the convergence to correct poses.

We evaluate our method on three widely-used benchmarks: the 3DMatch/3DLoMatch dataset~\cite{zeng20173dmatch,huang2021predator}, the ScanNet dataset~\cite{dai2017scannet}, and the ETH dataset~\cite{pomerleau2012challenging}.
With the help of the proposed sparse graph construction and IRLS with history reweighting, our method surpasses the current multiview registration baselines by $11.0\%$ and $6.2\%$ in registration recall on 3DMatch and 3DLoMatch, reduces the mean rotation and translation errors on ScanNet by $12.8\%$ and $13.8\%$. Meanwhile, our method shows strong generalization ability. Only trained on the indoor dataset, our method achieves a $99.8\%$ registration recall on the outdoor ETH dataset. Moreover, all the above state-of-the-art performances of our method only require $20\% \sim 40\%$ pairwise registrations of existing multiview point cloud registration methods, which demonstrates our computation efficiency.

%% file: content/02_relatedwork.tex
\begin{figure*}
\begin{center}
\includegraphics[width=\linewidth]{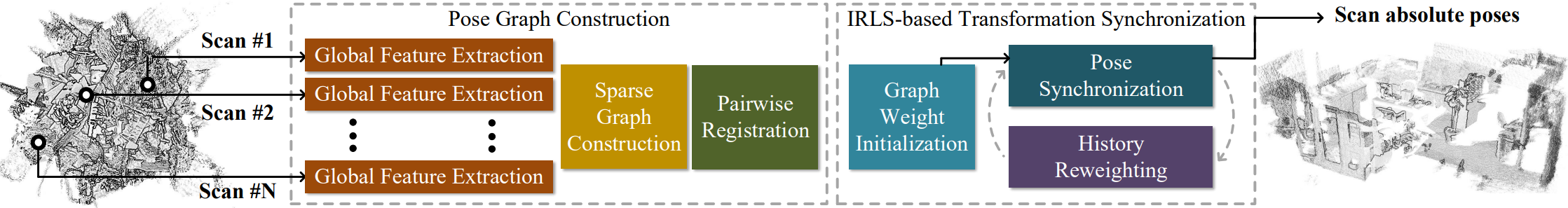}
\end{center}
\caption{The pipeline of the proposed method.}
\label{fig:pipeline}
\end{figure*}

\section{Related work}
\label{sec:rel}

\subsection{Pairwise registration}
There are mainly two kinds of pairwise point cloud registrations.
Feature-based methods extract a set of local descriptors~\cite{gojcic2019perfect,bai2020d3feat,choy2019fully,ao2021spinnet,wang2022you,wang2023roreg} on detected keypoints~\cite{bai2020d3feat,huang2021predator}. Then, local descriptors are matched to build correspondences~\cite{yu2021cofinet,qin2022geometric,li2022lepard,yew2022regtr}. Finally, correspondences are filtered~\cite{pais20203dregnet,choy2020deep,lee2021deep,bai2021pointdsc,chen2022sc2} and used in transformation estimation~\cite{wang2022you,qin2022geometric,yew2022regtr} to find rigid transformations.
Other works, known as direct registration methods, either directly regress the transformations~\cite{huang2020feature,aoki2019pointnetlk,li2020deterministic,yuan2020deepgmr} or refine correspondences~\cite{lu2019deepvcp,wang2019prnet,wang2019deep,fu2021robust} by considering the information from both point clouds with attention layers.
Our multiview registration method is based on the pairwise registration, which is compatible with all above methods.

\subsection{Multiview registration}
Most multiview point cloud registration methods~\cite{govindu2004lie,arrigoni2016spectral,huang2017translation,choi2015robust,birdal2018bayesian,huang2019learning,gojcic2020learning,yew2021learning,syncmatch} aim at recovering the absolute scan poses from exhaustive pairwise registrations.  
However, exhaustive pairwise registration is time-consuming~\cite{dong2018hierarchical} and may contain lots of outliers~\cite{yew2021learning}.
To reduce the computational burden, some traditional works~\cite{huber2003fully,mian2006three,guo2014accurate,dong2018hierarchical,wu2023hierarchical} resort to growing-based strategy to merge selected scans iteratively, which requires fewer pairwise registrations but may fail due to the accumulated errors in the growing process. In contrast, we avoid complex growing strategies to find inlier pairs but incorporate learning-based techniques to select reliable scan pairs, which enables more accurate subsequent synchronization.
Other works~\cite{huber2003fully,zhou2016fast,bhattacharya2019efficient,theiler2015globally,torsello2011multiview,choi2015robust,arrigoni2016spectral,huang2019learning,gojcic2020learning,yew2021learning} focus on pruning outliers on the constructed graph. IRLS-based scheme is one of the most prevalent technique~\cite{holland1977robust,arrigoni2016spectral,huang2017translation,bernard2015solution,huang2019learning,gojcic2020learning,yew2021learning}.
However, the iterative refinement of IRLS can easily trapped in a local minima and fails to prune out outlier edges~\cite{arrigoni2016spectral,yew2021learning}. 
The reweighting function is proved to be the most important design in a reliable IRLS~\cite{holland1977robust, arrigoni2016spectral,huang2019learning}.
Thus, recent learning-based advances~\cite{huang2019learning,gojcic2020learning,yew2021learning} adopt data-driven strategy to learn robust reweighting functions, which achieve impressive performances but cannot generalize well to unfamiliar graphs. We design a history reweighting function with strong generalization ability and robustness to outliers.

%% file: content/03_method.tex
\section{Method}
\label{sec:mth}

\subsection{Overview}
Consider a set of unaligned scans $\mathcal{P} = \{P_i| i=1,...,N\}$ in the same 3D scene. The target of multiview registration is to recover the underlying global scan poses $\{T_i = (R_i,t_i)\in SE(3)| i=1,...,N\}$. 
In the following, we first introduce how to initialize a pose graph with reliable edges in Sec.~\ref{sec:sparse}. 
Then, we propose a novel history reweighting function within a IRLS scheme in Sec.~\ref{sec:irls} to solve for the poses of every scan.
The pipeline is illustrated in Fig.~\ref{fig:pipeline}.

\subsection{Learn to construct a sparse graph}
\label{sec:sparse}

In this section, we aim to construct a pose graph for the multiview registration. Specifically, the graph is denoted by $\mathcal{G}(\mathcal{V}, \mathcal{E})$, where each vertex $v_i \in \mathcal{V}$ represents each scan $P_i$ while edge $(i,j) \in \mathcal{E}$ encodes the relative poses between scan $P_j$ and scan $P_i$. We will first estimate an overlap score $s_{ij}$ for each scan pair $(P_i,P_j)$. Then, given the overlap scores, we construct a sparse graph by selecting a set of scan pairs with large estimated overlaps and apply pairwise transformations on them only.

\textbf{Global feature extraction}. To extract the global feature $F$ for a point cloud $P$, we first 
downsample $P$ by voxels and 
extract a local feature $f_p \in \mathbb{R}^{d}$ on every sampled point $p \in P$ from its local 3D patch $N_p = \{p'|\|p-p'\|_2<r,p'\in P\}$ within a radius of $r$ by
\begin{equation}
    f_p = \varphi (N_p),
    \label{eq:local}
\end{equation}
where $\varphi$ is a neural network for extracting local descriptors, such as PointNet~\cite{qi2017pointnet}, FCGF~\cite{choy2019fully}, and YOHO~\cite{wang2022you}. 
By default, we adopt YOHO as the local descriptor~\cite{wang2022you} due to its superior performance.
Then, we apply a NetVLAD~\cite{arandjelovic2018netvlad} layer on the local features to extract a global feature $F$
\begin{equation}
    F = NetVLAD(\{f_p\}).
    \label{eq:global}
\end{equation}
Note $F\in \mathbb{R}^n$ is normalized such that $\|F\|_2=1$.

\textbf{Sparse graph construction}. For a scan pair $(P_i,P_j)$, we estimate their overlap score by
\begin{equation}
    s_{ij} = (F_i^{T}F_j+1)/2,
    \label{eq:sim}
\end{equation}
where $s_{ij} \in [0,1]$ indicates the overlap between $P_i$ and $P_j$. 
We train the $NetVLAD$ with a L1 loss between the predicted overlap score and the ground-truth overlap ratio.

For each scan, we select other \textit{k} scan pairs with the largest overlap scores to connect with the scan. This leads to a sparse graph with edges
\begin{equation}
    \mathcal{E}=\{(i,j:\argtopk_{P_j \in \mathcal{P}, j \neq i} s_{ij}), \forall P_i \in \mathcal{P}\}.
    \label{eq:edges}
\end{equation}

On each edge $(i,j) \in \mathcal{E}$ of the constructed graph, we estimate a relative pose $T_{ij}$ on the scan pair from their extracted local descriptors. By default, we follow \cite{wang2022you} to apply nearest neighborhood matcher on the local descriptors and estimate the relative pose from the RANSAC variant.

\textbf{Discussion}. Recent multiview registration methods~\cite{zhou2016fast,huang2019learning,gojcic2020learning,yew2021learning} usually exhaustively estimate all $\binom{N}{2}$ relative poses and many of these scan pairs have no overlap at all. In our method, we extract global features to determine the overlap scores to select $N \times k$ scan pairs. Actually, we only need to conduct pairwise registration less than $N \times k$ because the graph is an indirection graph and we only need to count each edge once.
Our global feature extraction is much more efficient than matching descriptors and running RANSAC in the pairwise registration. 
The subsequent pose synchronization only needs to operate on these sparse edges, which also improves the efficiency. 
Moreover, the retained pose graph contains much less fewer outliers, which thus improves the accuracy of the subsequent synchronization.

\subsection{IRLS with history reweighting}
\label{sec:irls}
In this section, we apply the Iteratively Reweighted Least Squares (IRLS) scheme to estimate the consistent global poses on all scans. The key idea of IRLS is to associate a weight on each edge to indicate the reliability of each scan pair. These weights are iteratively refined such that outlier edges will have small weights so these outlier relative poses will not affect the final global poses. In the following, we first initialize edge weights, and iteratively estimate poses based on edge weights and update edge weights with the proposed history reweighting function.



\subsubsection{Weight initialization}
\label{sec:init}
The weight $w^{(0)}_{ij}$ is initialized from both the estimated overlap score $s_{ij}$ and the quality of the pairwise registration by
\begin{equation}
    w^{(0)}_{ij} = s_{ij}*r_{ij},
\end{equation}
where $r_{ij}$ reveals the quality of pairwise registration. In the pairwise registration, a set of correspondences $C=\{(p,q)|p \in P_i, q \in P_j\}$ are established by matching local descriptors. Thus, $r_{ij}$ is defined as the number of inlier correspondences in $C$ conforming with $T_{ij} = (R_{ij},t_{ij})$, which is
\begin{equation}
    r_{ij} = \sum_{(p,q) \in C} [\![ \|p-R_{ij}q-t_{ij}\|^2 < \tau]\!],
    \label{eq:inlier}
\end{equation}
where $[\![ \cdot ]\!]$ is the Iverson bracket, $\tau$ is a pre-defined inlier threshold.



\subsubsection{Pose synchronization}
\label{sec:syn}
Given the edge weights and input relative poses $\{w_{ij}, T_{ij}=(R_{ij},t_{ij})|(i,j) \in \mathcal{E}\}$, we solve for the global scan poses $\{T_i=(R_i,t_i)\}$. 
We adopt the closed-form synchronization algorithm proposed in \cite{arie2012global,huang2019learning}.
We first compute the rotations by rotation synchronization~\cite{arie2012global,gojcic2020learning}, and then compute the translations by translation synchronization~\cite{huang2019learning}. 

\textbf{Rotation synchronization}. 
The goal of rotation synchronization is to solve
\begin{equation}
    \{R_1,...R_N\} = \mathop{\arg\min}_{R_1,...R_N \in SO(3)} \sum_{(i,j) \in \mathcal{E}} w_{ij}\|R_{ij} - R_i^TR_j\|^2_F,
    \label{eq:rotsyn}
\end{equation}
where $\|\cdot\|_F$ means the Frobenius norm of the matrix.
The problem has a closed-form solution, which can be derived from the eigenvectors of a symmetric matrix $L \in \mathbb{R}^{3N*3N}$
\begin{equation}\small
L = 
\left(              
  \begin{array}{cccc}  
    \sum\limits_{(1,j) \in \mathcal{E}}{w_{1j}} \mathbf{I}_3 & -w_{12}R_{12} & \cdots &  -w_{1N}R_{1N} \\ 
    -w_{21}R_{21} & \sum\limits_{(2,j) \in \mathcal{E}}{w_{2j}} \mathbf{I}_3 & \cdots &  -w_{2N}R_{2N} \\ 
    \vdots  & \vdots & \ddots &  \vdots \\ 
    -w_{N1}R_{N1} & -w_{N2}R_{N2} & \cdots   &  \sum\limits_{(N,j) \in \mathcal{E}}{w_{Nj}} \mathbf{I}_3\\
  \end{array}
\right) 
\end{equation}
$L$ is a sparse matrix since the constructed graph is sparse.
Given three eigenvectors $\tau_1 , \tau_2, \tau_3 \in \mathbb{R}^{3N}$ corresponding to the three smallest eigenvalues $\lambda_1 < \lambda_2 < \lambda_3$ of $L$, we stack these three eigenvectors to construct a matrix $V = [\tau_1,\tau_2,\tau_3] \in \mathbb{R}^{3N*3}$.
Then, $R_i$ can be derived by projecting $v_i = V[3i-3:3i] \in \mathbb{R}^{3*3}$ to $SO(3)$. More details can be found in the supplementary material.

\textbf{Translation synchronization}.
Similarly, translation synchronization retrieves the translation vectors $\{t_i\}$ that minimize the problem:
\begin{equation}
    \{t_1,...,t_N\} = \mathop{\arg\min}_{t_1,...,t_N \in \mathbb{R}^3} \sum_{(i,j) \in \mathcal{E}} w_{ij}\|R_{i}t_{ij}+t_i-t_{j}\|^2
    \label{eq:base_transsyn}
\end{equation}
We solve it by the standard least square method~\cite{huang2019learning}.

\subsubsection{History reweighting function}
\label{sec:reweight}
Given the synchronized poses, we re-compute weights on edges such that outlier edges will have smaller weights than the inlier edges. 
Assume the synchronized poses at the $n$-th iteration are $\{T_i^{(n)}=(R_i^{(n)},t_i^{(n)})\}$. We first compute the rotation residual $\delta^{(n)}_{ij}$ by
\begin{equation}
    \delta^{(n)}_{ij} = \Delta(R_{ij}, R_i^{(n)T} R_j^{(n)}),
\end{equation}
where $\Delta(R_1,R_2)$ means the angular difference between the rotation $R_1$ and $R_2$. $\Delta(R_1,R_2)$ is implemented by transforming $R_1^{T}R_2$ into an axis-angle form and outputing the rotation angle. Then, the updated weights are computed from rotation residuals of all previous iterations by
\begin{equation}
    w^{(n)}_{ij}=w^{(0)}_{ij}\exp\left(-\sum_{m=1}^n{g(m)\delta_{i,j}^{(m)}}\right),
    \label{eq:reweight}
\end{equation}
where $g(m)$ is a predefined coefficient function of the iteration number with $g(m)>0$. We will elaborate the design of $g(m)$ later. Instead, we first discuss the intuition behind the Eq.~(\ref{eq:reweight}).

\textbf{Intuition of Eq.~(\ref{eq:reweight})}. Similar to previous reweighting functions~\cite{arrigoni2016spectral,yew2021learning,gojcic2020learning}, a larger rotation residual $\delta$ will lead to a smaller weight because large residuals are often caused by outliers. Meanwhile, there are two differences from previous reweighting functions. First, we multiply the initial weights $w^{(0)}_{ij}$ so that the recomputed weights always retain information from the warm-start initialization in Sec.~\ref{sec:init} and these initialized weights will be adjusted by the residuals in the iterative refinement. Second, the weight at a specific iteration $n$ considers the residuals of all previous iterations $m\le n$. This design is inspired from the momentum optimization method RMSProp or Adam~\cite{kingma2014adam}, which utilizes the gradients in the history to stabilize the optimization process. Here, we adopt similar strategy to consider all residuals in the history to determine a robust weight for the current iteration, which is less sensitive to outliers.

\textbf{Design of coefficient function $g(m)$}. $g(m)$ can be regarded as a weight function. A small value of $g(m)$ means that we do not trust the residual at the iteration $m$ and this residual may not correctly identify inliers and outliers. In our observation, the residuals estimated by the first few iterations are not very stable so we want $g(m)$ is increasing with the iteration number $m$. Meanwhile, if we want to conduct $M$ IRLS iterations in total, we want the sum of coefficients at the final iteration $M$ will be 1, i.e. $\sum_{m=1}^M g(m)=1$. Thus, in our design, we have
\begin{equation}
    g(m)=\frac{2m}{M(M+1)}.
\end{equation}

After computing the updated weights, we iteratively synchronize the poses with these updated weights as stated in Sec.~\ref{sec:syn} and compute new weights from these new poses as stated in Sec.~\ref{sec:reweight}. The IRLS run $M$ iterations in total and the synchronized poses at the final iteration are regarded as the output poses for all scans.

%% file: content/04_experiments.tex
\section{Experiments}
\label{sec:exp}

\begin{figure*}
\begin{center}
\includegraphics[width=\linewidth]{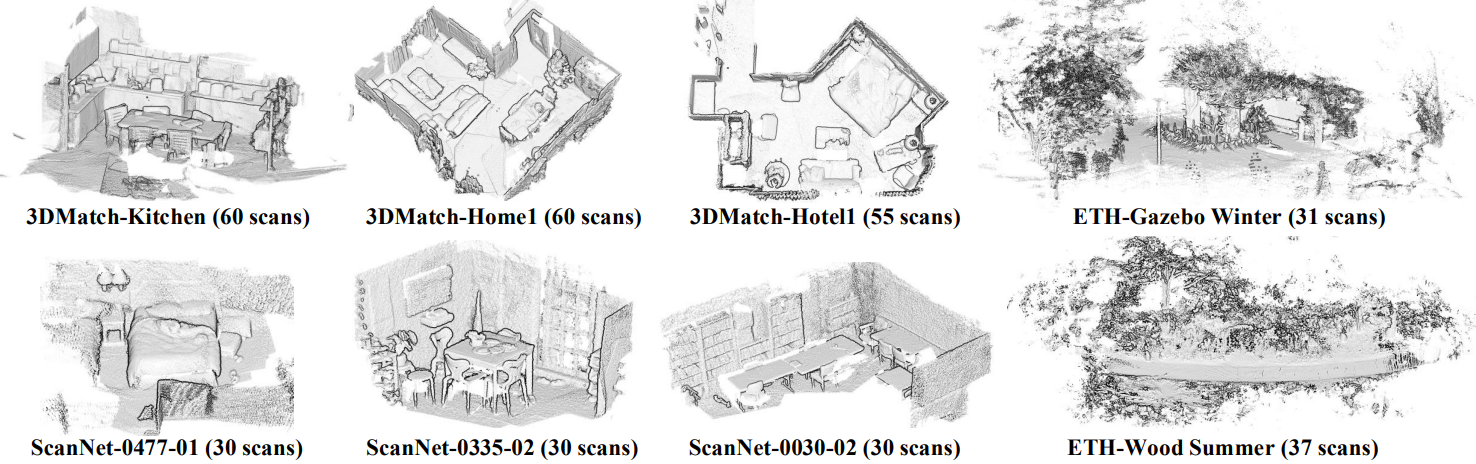}
\end{center}
\vspace{-13pt}
\caption{Qualitative results on the 3DMatch, ScanNet, and ETH datasets.}
\vspace{-10pt}
\label{fig:visual}
\end{figure*}

\subsection{Experimental protocol}
\subsubsection{Datasets}
\label{sec:exp_dataset}
We evaluate the proposed method on three widely used datasets:  3D(Lo)Match~\cite{zeng20173dmatch,huang2021predator}, ScanNet~\cite{dai2017scannet}, and ETH~\cite{pomerleau2012challenging} as follows.

\textbf{3DMatch} contains scans collected from 62 indoor scenes among which 46 are split for training, 8 for validation, and 8 for testing. Each test scene contains 54 scans on average. We follow previous works~\cite{gojcic2020learning,huang2021predator} to use 1623 scan pairs with $>30\%$ overlap ratio and 1781 scan pairs with $10\% \sim 30\%$ overlap as two test sets, denoted as 3DMatch and 3DLoMatch, respectively.

\textbf{ScanNet} contains RGBD sequences of 1513 indoor scenes. We follow~\cite{gojcic2020learning} to use the same 32 test scenes and convert 30 RGBD images that are 20 frames apart to 30 scans on each scene. 
There are 960 scans in total and we exhaustively select all 13920 scan pairs for evaluation.

\textbf{ETH} has 4 outdoor scenes with large domain gaps to the 3DMatch dataset and each scene contains 33 scans on average. 713 scan pairs are officially selected for evaluation. 

Our model is only trained on the training split of 3DMatch and evaluated on 3D(Lo)Match, ScanNet, and ETH. More training details can be found in supplementary material. 
For evaluation, we first perform multiview registration to recover the global scan poses. Then, we follow~\cite{gojcic2020learning} to evaluate the multiview registration quality on pairwise relative poses computed from the recovered global poses.
By default, we set $k$ in sparse graph construction to 10 for two indoor datasets and 6 for the ETH dataset.

\subsubsection{Metrics}
We follow~\cite{choy2019fully,ao2021spinnet,wang2022you,qin2022geometric} to adopt Registration Recall (RR) for evaluation on 3D(Lo)Match and ETH. RR reports the ratio of correctly aligned scan pairs.
A scan pair is regarded correctly-aligned if the average distance between the points under the estimated transformation $(R_{pre},t_{pre})$ and these points under the ground truth transformation $(R_{gt},t_{gt})$ is less than 0.2m for the 3D(Lo)Match dataset and 0.5m for the ETH dataset.
RR of all methods is calculated on the same official evaluation scan pairs mentioned in Sec.~\ref{sec:exp_dataset}.

For the evaluation on ScanNet, we follow~\cite{huang2019learning,gojcic2020learning,yew2021learning} to report Empirical Cumulative Distribution Functions (ECDF) of the rotation error $re$ and translation error $te$:
\begin{equation}
\label{eq:rete}
re = arccos\left(\frac{tr(R_{pre}^TR_{gt})-1}{2}\right)\ \ te = \|t_{pre}-t_{gt}\|^2.
\end{equation}

We also report the number of required pairwise registrations to initialize the pose graphs, denoted as ``\#Pair''.

\subsubsection{Baselines}
We compare the proposed method against several multiview registration baselines: EIGSE3~\cite{arrigoni2016spectral}, L1-IRLS~\cite{chatterjee2017robust}, RotAvg~\cite{chatterjee2017robust}, LMVR~\cite{gojcic2020learning}, LITS~\cite{yew2021learning}, and HARA~\cite{lee2022hara}. 
Specifically, LMVR is an end-to-end method, which performs pairwise registration and transformation synchronization in a single deep neural network. 
EIGSE3 proposes a spectral approach to solve transformation synchronization and further applies IRLS with Cauchy~\cite{holland1977robust} reweighting function to improve robustness.
L1-IRLS and RotAvg are two robust algorithms, which perform IRLS-based rotation synchronization using $l_1$ and $l_{1/2}$ reweighting functions to resist outliers.
HARA is a state-of-the-art handcrafted synchronization method, which conducts growing-based edge pruning by checking cycle consistency and performs IRLS-based synchronization using $l_{1/2}$ reweighting functions on the retained edges.
LITS is a state-of-the-art learning-based transformation synchronization method.
All baseline methods except LMVR are compatible with any pairwise registration methods~\cite{choy2019fully,ao2021spinnet,wang2022you,qin2022geometric}. Thus, we compare our method with these baselines using different pairwise registration algorithms, including FCGF~\cite{choy2019fully}, SpinNet~\cite{ao2021spinnet}, YOHO~\cite{wang2022you}, GeoTransformer~\cite{qin2022geometric}.

\subsubsection{Pose graph construction}
For a fair comparison with baseline multiview registration methods, we report the performances produced on three different types of input pose graphs. The first type ``Full" does not prune any edge so the pose graph is fully-connected. The second type ``Pruned" prunes edges according to the quality of pairwise registration, which is adopted by previous methods LITS~\cite{yew2021learning} and LMVR~\cite{gojcic2020learning} (called ``Good" in their papers). ``Pruned" first applies pairwise registration algorithms (FCGF~\cite{choy2019fully}, YOHO~\cite{wang2022you}, SpinNet~\cite{ao2021spinnet} or GeoTransformer~\cite{qin2022geometric}) to exhaustively register all scan pairs and then only retain scan pairs whose median point distance in the registered overlapping region is less than 0.05m~\cite{gojcic2020learning,yew2021learning} (0.15m for ETH).
The final type ``Ours" applies the proposed global feature for the overlap score estimation and constructs a sparse graph according to scores.

\begin{table}
\begin{center}
\resizebox{\linewidth}{!}{
\begin{tabular}{l|l|c|c|c|c}
\toprule[0.35mm]
\multicolumn{1}{c|}{\textit{Pose}}                                     & \multicolumn{1}{c|}{\multirow{2}{*}{\textit{Method}}} & \multicolumn{1}{c|}{\multirow{2}{*}{\textit{\#Pair}}} & \multicolumn{1}{c|}{SpinNet~\cite{ao2021spinnet}}                & \multicolumn{1}{c|}{YOHO~\cite{wang2022you}}                   & GeoTrans~\cite{qin2022geometric}               \\
\multicolumn{1}{c|}{\textit{Graph}}                                    & \multicolumn{1}{c|}{}                                 & \multicolumn{1}{c|}{}                                     & \multicolumn{1}{c|}{\textit{3D / 3DL-RR (\%)}} & \multicolumn{1}{c|}{\textit{3D / 3DL-RR (\%)}} & \textit{3D / 3DL-RR (\%)} \\\hline
\hline
\multirow{6}{*}{Full}                                                  & EIGSE3~\cite{arrigoni2016spectral}                                                & 11905                                                     & 20.8  /  13.6                                 & 23.2  /  6.6                                  & 17.0  /  9.1             \\
                                                                       & L1-IRLS~\cite{chatterjee2017robust}                                               & 11905                                                     & 49.8  /  29.4                                 & 52.2  /  32.2                                 & 55.7  /  37.3            \\
                                                                       & RotAvg~\cite{chatterjee2017robust}                                                & 11905                                                     & 59.3  /  38.9                                 & 61.8  /  44.1                                 & 68.6  /  56.5            \\
                                                                       & LITS~\cite{yew2021learning}                                                  & 11905                                                     & 68.1  /  47.9                                 & 77.0  /  59.0                                 & 84.2  /  73.0            \\
                                                                       & HARA~\cite{lee2022hara}      &    11905                & 82.7 / 63.6          & 83.1 / 68.7          & 83.4 / 68.5             \\
                                                                       & Ours                                                  & 11905                                                     & 93.3  /  77.2                                 & 93.2  /  76.8                                 & 91.5  /  82.4            \\\cdashline{1-6}[3pt/3pt]
\multirow{6}{*}{\begin{tabular}[c]{@{}l@{}}Pruned\\\cite{gojcic2020learning}\end{tabular}} & EIGSE3~\cite{arrigoni2016spectral}                                                & 11905                                                     & 42.7  /  34.6                                 & 40.1  /  26.5                                 & 39.4  /  28.7            \\
                                                                       & L1-IRLS~\cite{chatterjee2017robust}                                               & 11905                                                     & 66.9  /  46.2                                 & 68.6  /  49.0                                 & 77.4  /  58.3            \\
                                                                       & RotAvg~\cite{chatterjee2017robust}                                                & 11905                                                     & 72.8  /  55.3                                 & 77.2  /  60.3                                 & 81.6  /  68.5            \\
                                                                       & LITS~\cite{yew2021learning}                                                  & 11905                                                     & 73.1  /  55.5                                 & 80.8  /  65.2                                 & 84.6  /  76.8            \\
                                                                       & HARA~\cite{lee2022hara}    &  11905                & 84.0 / 62.5          & 83.8 / 71.9          & 84.9 / 73.7              \\
                                                                       & Ours                                                  & 11905                                                     & 94.8  /  \textbf{80.6}                                 & 95.2  /  \textbf{82.3}                                 & 95.2  /  82.8            \\\cdashline{1-6}[3pt/3pt]
Ours                                                                   & Ours                                         & \textbf{2798}                                                      & \textbf{94.9}  /  80.0                                     & \textbf{96.2}  /  81.6                                 & \textbf{95.9} / \textbf{83.0}                \\ 
\bottomrule[0.35mm]
\end{tabular}}
\end{center}
\vspace{-15pt}
\caption{Registration recall on the 3DMatch (``3D") and 3DLoMatch (``3DL") datasets. 
We report results with different pairwise registration algorithms (SpinNet~\cite{ao2021spinnet}, YOHO~\cite{wang2022you}, GeoTrans~\cite{qin2022geometric}).}
\vspace{-10pt}
\label{tab:3dmatch}
\end{table}

\begin{table*}\footnotesize
\begin{center}
\resizebox{\linewidth}{!}{
\begin{tabular}{l|l|c|cccccc|cccccc}
\toprule[0.25mm]
\multirow{2}{*}{\textit{Pose Graph}}                                               & \multicolumn{1}{c|}{\multirow{2}{*}{\textit{Method}}}  & \multicolumn{1}{c|}{\multirow{2}{*}{\textit{\#Pair}}} & \multicolumn{6}{c|}{\textit{Rotation   Error}}                                                                & \multicolumn{6}{c}{\textit{Translation   Error (m)}}                                               \\
                                                                          & \multicolumn{1}{c|}{}        & \multicolumn{1}{c|}{}                          & 3°            & 5°            & 10°           & 30°           & 45°           & \multicolumn{1}{c|}{Mean/Med} & 0.05          & 0.1           & 0.25          & 0.5           & 0.75          & Mean/Med           \\\hline
                                                                          \hline
\multirow{7}{*}{Full}       & LMVR~\cite{gojcic2020learning}                & 13920                 & 48.3          & 53.6          & 58.9          & 63.2          & 64.0          & 48.1°/33.7°                   & 34.5          & 49.1          & 58.5          & 61.6          & 63.9          & 0.83/0.55          \\
                            & LITS~\cite{yew2021learning}                   & 13920                                  & 47.4          & 58.4          & 70.5          & 78.3          & 79.7          & 27.6°/-                       & 29.6          & 47.5          & 66.7          & 73.3          & 77.6          & 0.56/-             \\
                            & EIGSE3~\cite{arrigoni2016spectral}*           & 13920                                       & 19.7          & 24.4          & 32.3          & 49.3          & 56.9          & 53.6°/48.0°                   & 11.2          & 19.7          & 30.5          & 45.7          & 56.7          & 1.03/0.94          \\
                            & L1-IRLS~\cite{chatterjee2017robust}*          & 13920                                       & 38.1          & 44.2          & 48.8          & 55.7          & 56.5          & 53.9°/47.1°                   & 18.5          & 30.4          & 40.7          & 47.8          & 54.4          & 1.14/1.07          \\
                            & RotAvg~\cite{chatterjee2017robust}*           & 13920                                       & 44.1          & 49.8          & 52.8          & 56.5          & 57.3          & 53.1°/44.0°                   & 28.2          & 40.8          & 48.6          & 51.9          & 56.1          & 1.13/1.05          \\
                            & LITS~\cite{yew2021learning}*                  & 13920                                  & 52.8          & 67.1          & 74.9          & 77.9          & 79.5          & 26.8°/27.9°                   & 29.4          & 51.1          & 68.9          & 75.0          & 77.0          & 0.68/0.66          \\
                           & HARA~\cite{lee2022hara}* & 13920                & 54.9                 & 64.3                 & 71.3                 & 74.1                 & 74.2                 & 32.1°/29.2°          & 35.8                 & 54.4                 & 66.3                 & 69.7                 & 72.9                 & 0.87/0.75  \\          
                            & Ours                                   & 13920           & 57.2          & 68.5          & 75.1          & 78.1          & 78.8          & 26.4°/19.5°                   & 39.4          & 61.5          & 72.0          & 75.2          & 77.6          & 0.70/0.59          \\\cdashline{1-15}[3pt/3pt]
\multirow{5}{*}{Pruned~\cite{gojcic2020learning}}    & EIGSE3~\cite{arrigoni2016spectral}*             & 13920                             & 40.8          & 46.3          & 51.9          & 61.2          & 65.7          & 40.6°/37.1°                   & 23.9          & 38.5          & 51.0          & 59.3          & 66.1          & 0.88/0.84          \\
                            & L1-IRLS~\cite{chatterjee2017robust}*          & 13920                              & 46.3          & 54.2          & 61.6          & 64.3          & 66.8          & 41.8°/34.0°                   & 24.1          & 38.5          & 48.3          & 55.6          & 60.9          & 1.05/1.01          \\
                            & RotAvg~\cite{chatterjee2017robust}*           & 13920                                 & 50.2          & 60.1          & 65.3          & 66.8          & 68.8          & 38.5°/31.6°                   & 31.8          & 49.0          & 58.8          & 63.3          & 65.6          & 0.96/0.83          \\
                            & LITS~\cite{yew2021learning}*                 & 13920                             & 54.3          & 69.4          & 75.6          & 78.5          & 80.3          & 24.9°/19.9°                   & 31.4          & 54.4          & 72.3          & 76.7          & 79.6          & 0.65/0.56          \\
                           & HARA~\cite{lee2022hara}* & 13920                & 55.7                 & 63.7                 & 69.0                 & 70.8                 & 72.1                 & 34.7°/31.3°          & 35.2                 & 53.6                 & 65.4                 & 68.6                 & 71.7                 & 0.86/0.71            \\
                            & Ours                                        & 13920      & \textbf{59.4}         & 71.9          & 80.0          & 82.1          & 82.6          & \textbf{21.7°}/19.1°                   & \textbf{39.9}          & 63.0          & 74.3          & 77.6          & 80.2          & 0.64/\textbf{0.47}          \\ \cdashline{1-15}[3pt/3pt]
Ours                      & Ours                              & \textbf{6004}             & 59.1 & \textbf{73.1} & \textbf{80.8} & \textbf{82.5} & \textbf{83.0} & \textbf{21.7°/19.0°}          & \textbf{39.9} & \textbf{64.1} & \textbf{76.7} & \textbf{79.0} & \textbf{81.9} & \textbf{0.56}/0.49\\
\bottomrule[0.25mm]
\end{tabular}}
* means using the same selected frames and pairwise transformations as ours.
\vspace{-10pt}
\end{center}
\caption{Registration performance on the ScanNet dataset. The pairwise registration algorithm for all methods is YOHO~\cite{wang2022you} except for LMVR~\cite{gojcic2020learning} which includes pairwise registration in its pipeline.}
\vspace{-5pt}
\label{tab:scannet}
\end{table*}

\begin{table}
\begin{center}
\resizebox{\linewidth}{!}{
\begin{tabular}{l|l|c|ccc}
\toprule[0.35mm]
\multicolumn{1}{c|}{\multirow{2}{*}{\textit{Pose Graph}}} & \multicolumn{1}{c|}{\multirow{2}{*}{\textit{Method}}}  & \multicolumn{1}{c|}{\multirow{2}{*}{\textit{\#Pair}}} & FCGF~\cite{choy2019fully}             & SpinNet~\cite{ao2021spinnet}           & YOHO~\cite{wang2022you}              \\
\multicolumn{1}{c|}{}                                     & \multicolumn{1}{c|}{}                & \multicolumn{1}{c|}{}                                            & \textit{RR (\%)} & \textit{RR (\%)} & \textit{RR (\%)} \\\hline
\hline
\multirow{5}{*}{Full}     & EIGSE3~\cite{arrigoni2016spectral}                   &2123                                   & 44.8                 & 56.3                    & 60.9                 \\
                          & L1-IRLS~\cite{chatterjee2017robust}                  &2123                                    & 60.5                 & 73.2                    & 77.2                 \\
                          & RotAvg~\cite{chatterjee2017robust}                   &2123                                    & 67.3                 & 82.1                    & 85.4                 \\
                          & LITS~\cite{yew2021learning}                          &2123                               & 26.3                 & 36.4                    & 34.8                 \\
                          & HARA~\cite{lee2022hara}                          &2123                               & 72.2                 & 79.3                    & 85.4                 \\
                          & Ours                                                 &2123        & 85.7                 & 86.3                    & 98.8                 \\\cdashline{1-6}[3pt/3pt]
\multirow{5}{*}{Pruned~\cite{gojcic2020learning}} & EIGSE3~\cite{arrigoni2016spectral}              &2123                                         & 89.4                 & 93.6                    & 96.3                 \\
                          & L1-IRLS~\cite{chatterjee2017robust}                  &2123                                    & 86.1                 & 87.9                    & 90.2                 \\
                          & RotAvg~\cite{chatterjee2017robust}                   &2123                                    & 95.6                 & 95.5                    & 96.6                 \\
                          & LITS~\cite{yew2021learning}                          &2123                               & 41.2                 & 47.3                    & 48.4                 \\
                          & HARA~\cite{lee2022hara}                          &2123                               & 90.3                 & 97.8                    & 96.0                 \\
                          & Ours                                                 &2123        & 96.8                 & \textbf{99.8}           & 97.2                 \\\cdashline{1-6}[3pt/3pt]
\multicolumn{1}{l|}{Ours}  & Ours                                      &\textbf{516}         & \textbf{97.4}        & \textbf{99.8}           & \textbf{99.1}        \\ 
\bottomrule[0.35mm]
\end{tabular}}
\vspace{-10pt}
\end{center}
\caption{Registration recall on the ETH dataset. We report results using different pairwise registration algorithms (FCGF~\cite{choy2019fully}, SpinNet~\cite{ao2021spinnet}, 
YOHO~\cite{wang2022you}).}
\label{tab:eth}
\end{table}

\subsection{Results on three benchmarks}
Qualitative results are shown in Fig.~\ref{fig:visual}. Quantitative results on the 3DMatch, the ScanNet and the ETH datasets are shown in Table~\ref{tab:3dmatch}, Table~\ref{tab:scannet} and Table~\ref{tab:eth}, respectively. 

First, the results show that our method achieves significant better performances than all baseline methods with $\sim$5\%-10\% improvements on the 3DMatch and 3DLoMatch dataset, which demonstrates that our method is able to accurately align low-overlapped scan pairs via pose synchronization. Meanwhile, our method only requires $\sim 30\%$ pairwise registrations with the help of our sparse graph construction, which greatly improves the efficiency. 

Second, when using the same pose graphs as previous method, our method already achieves better performances on all datasets, which is benefited from our history reweighting function in the IRLS. Meanwhile, applying our global features for the graph construction further improves the results, which demonstrates the predicted overlap score is more robust than simply pruning edges according to the pairwise registration.

Finally, the results on the outdoor ETH dataset demonstrate the generalization ability of the proposed method. Both our method and the learning-based method LITS~\cite{yew2021learning} is trained on the indoor 3DMatch dataset, However, LITS does not generalize well to outdoor dataset (only $\sim$45\% recall) even though it shows strong performances on both indoor datasets. In comparison, our method still achieves strong performances (almost 100\% registration recall) on the outdoor dataset.

\subsection{Analysis}
We thoroughly conduct analyses on the proposed designs about the pose graph construction and history reweighting IRLS modules in this section. By default, all analyses are conducted on the 3D(Lo)Match dataset with YOHO~\cite{wang2022you} as the pairwise registration method.

\subsubsection{Sparse Graph Construction}

\begin{figure}
\begin{center}
\includegraphics[width=1\linewidth]{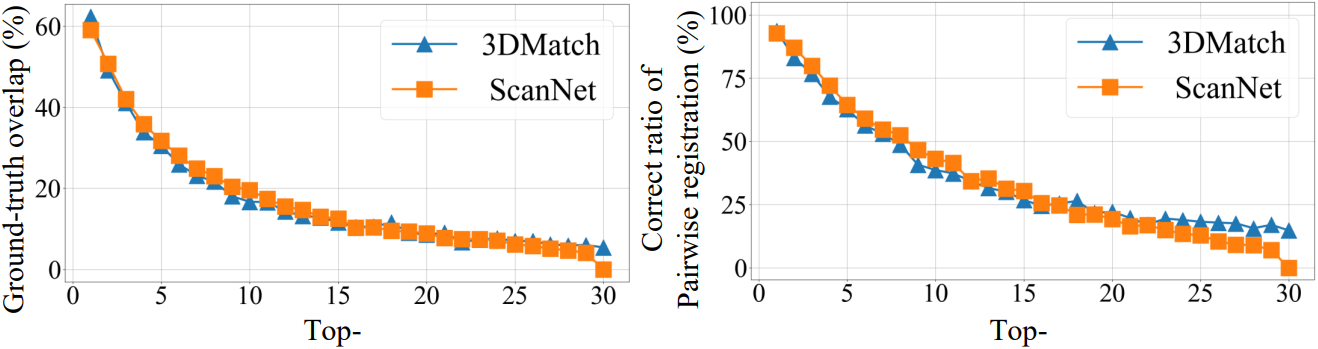}
\end{center}
\vspace{-10pt}
\caption{Ground truth overlap ratios and correct ratios of pairwise registration with Top-k overlap scores.}
\vspace{-10pt}
\label{fig:gtover}
\end{figure}

\begin{figure}
\begin{center}
\includegraphics[width=0.8\linewidth]{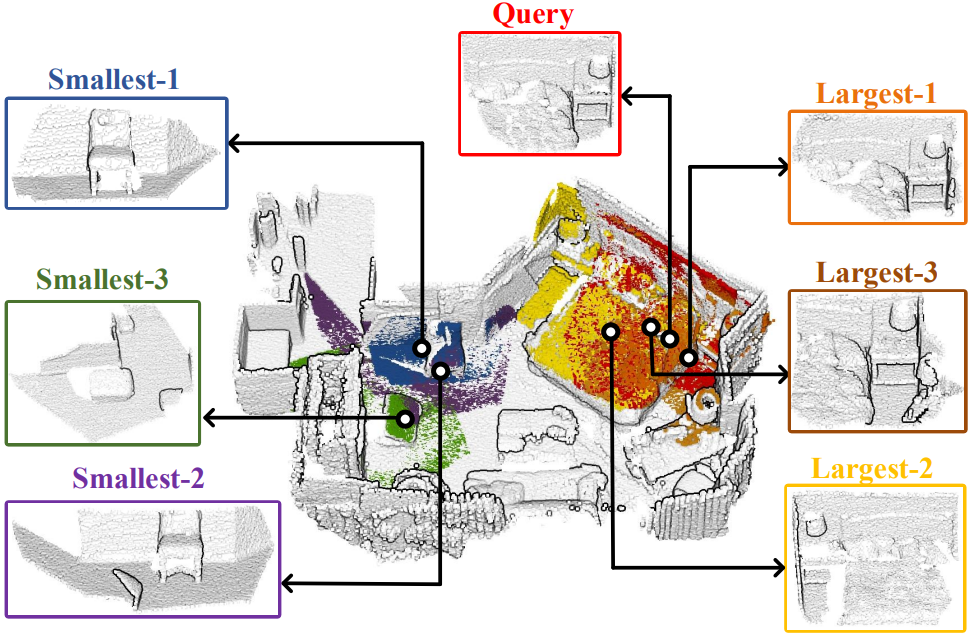}
\end{center}
\vspace{-10pt}
\caption{An example of retrieving scans using the global feature. The predicted top-3 scans with largest overlap scores indeed have large overlaps with the query scan while the 3 scans with smallest overlap scores are far away from the query.}
\vspace{-5pt}
\label{fig:simcase}
\end{figure}

\textbf{Are predicted overlap scores well-calibrated}? Well-calibrated overlap scores should assign higher scores to scan pairs with more overlap regions. 
Meanwhile, we want the scan pairs with higher overlap scores can be easily aligned by the pairwise registration algorithms.
In Fig.~\ref{fig:gtover}, we report the averaged ground truth overlap ratios and correct ratio of pairwise registration of scan pairs with top-30 predicted overlap scores. It can be seen that the estimated overlap scores are able to identify the reliable scan pairs with high overlap ratios. A visualization of retrieved scans using the global feature is given in Fig.~\ref{fig:simcase}.

\textbf{Can our sparse graphs improve other multiview registration methods}?
We compare the performance of EIGSE3~\cite{arrigoni2016spectral}, RotAvg~\cite{chatterjee2017robust}, and LITS~\cite{yew2021learning} using the fully-connected pose graph (``Full"), outliers pruned by pairwise registration results~\cite{huang2019learning,gojcic2020learning,yew2021learning} (``Pruned") and the proposed sparse graph (``Ours") in Table~\ref{tab:sparse}.
It can be seen that our sparse graph construction boosts the performance of baseline methods by a larger margin than ``Pruned" graphs. Note ``Pruned" requires exhaustive pairwise registration while we only need to conduct pairwise registration on the retained edges. Thus, our method is more efficient. Detailed running times are provided in the supplementary material.

\begin{table}\scriptsize
\begin{center}
\resizebox{\linewidth}{!}{
\begin{tabular}{l|l|c|cc}
\toprule[0.25mm]
\textit{Pose Graph} & \textit{Method}  & \textit{\#Pair} & \textit{3D-RR (\%)} & \textit{3DL-RR (\%)} \\\hline
\hline
Full                & EIGSE3~\cite{arrigoni2016spectral}     & 11905     & 23.2                & 6.6                    \\
Pruned~\cite{gojcic2020learning}            & EIGSE3~\cite{arrigoni2016spectral}   & 11905       & 40.1                & 26.5                   \\
Ours                & EIGSE3~\cite{arrigoni2016spectral}    & \textbf{2798}      & \textbf{60.4}                & \textbf{44.6}                  \\\cdashline{1-5}[3pt/3pt]
Full                & RotAvg~\cite{chatterjee2017robust}    & 11905      & 61.8                & 44.1                   \\
Pruned~\cite{gojcic2020learning}            & RotAvg~\cite{chatterjee2017robust}    & 11905      & 77.2                & 60.3                   \\
Ours                & RotAvg~\cite{chatterjee2017robust}     & \textbf{2798}     & \textbf{81.7}                & \textbf{63.9}                   \\\cdashline{1-5}[3pt/3pt]
Full                & LITS~\cite{yew2021learning}     & 11905       & 77.0                & 59.0                   \\
Pruned~\cite{gojcic2020learning}            & LITS~\cite{yew2021learning}    & 11905        & 80.8                & 65.2                   \\
Ours                & LITS~\cite{yew2021learning}     & \textbf{2798}       & \textbf{84.6}                & \textbf{68.6}                   \\
\bottomrule[0.25mm]
\end{tabular}}
\end{center}
\vspace{-10pt}
\caption{Performances of applying different multiview registration methods on different input pose graphs.}
\vspace{-5pt}
\label{tab:sparse}
\end{table}

\subsubsection{Ablation studies on history reweighting}
\label{sec:opt}
\begin{figure}
\begin{center}
\includegraphics[width=\linewidth]{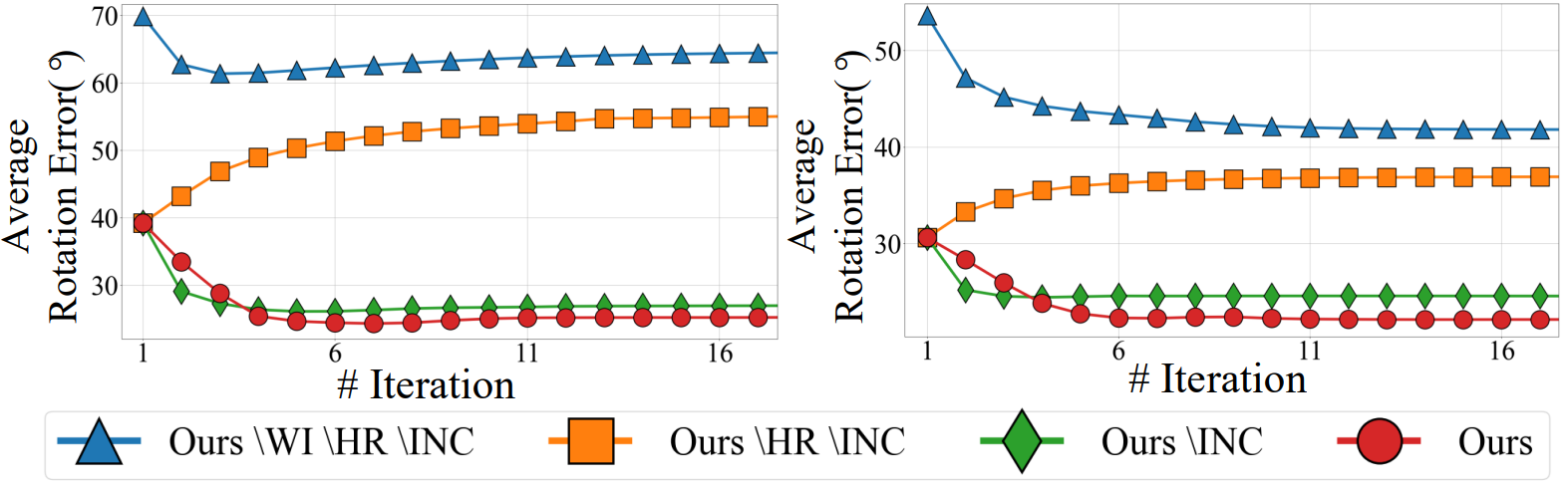}
\end{center}
\vspace{-10pt}
\caption{Curves of rotation error w.r.t. iteration number with ablation on specific components of our IRLS scheme on the 3DMatch (left) and  the ScanNet (right). ``\textbackslash'' means ``without''.}
\label{fig:irlsw}
\vspace{-10pt}
\end{figure}

We conduct ablation studies on our designs in the proposed IRLS algorithm. 
The results are shown in Table.~\ref{tab:ablsync} and the convergence curves are shown in Fig.~\ref{fig:irlsw}. 
We consider the following three designs.
1) \textit{Weight initialization} (\textit{WI}). We initialize the weight to be the product of both the inlier correspondence number $r_{ij}$ and the predicted overlap score $s_{ij}$. Alternatively, we may just initialize the weight with $r_{ij}$ or $s_{ij}$ only. Results show that the proposed initialization is better.
2) \textit{History reweighting} (\textit{HR}). In our reweighting function, the recomputed weight is determined by rotation residuals of all previous iterations. Alternatively, we may just compute the weight from the rotation residual of current iteration.
History reweighting stabilizes the iterative refinement and makes IRLS more robust to outliers.
3) \textit{Designing $g(m)$ to be increasing with $m$} (\textit{INC}). In our design, we set $g(m)$ to be increasing with $m$ so that the residuals at early iterations will have smaller impacts on results. Alternatively, we may set $g(m)=1/M$ so that all residuals contribute equally to the weights. However, rotations estimated in the early stage are not very stable so that reducing their impacts will improve the results.


\begin{table}
\begin{center}
\resizebox{\linewidth}{!}{
\begin{tabular}{c|cc|cc|c}
\toprule[0.35mm]
                      & \multicolumn{2}{c|}{\textit{Initialization}} & \multicolumn{2}{c|}{\textit{Reweighting}} & \multirow{2}{*}{\textit{Full}} \\
                      & \textit{w/o $s_{ij}$}      & \textit{w/o $r_{ij}$}     & \textit{w/o HR}     & \textit{w/o INC}    &                                \\\hline
                      \hline
\textit{3D-RR(\%)}   & 95.5 (-0.7)           & 76.9 (-19.3)         & 83.1 (-13.1)        & 94.1 (-2.1)         & 96.2                           \\
\textit{3DL-RR(\%)} & 79.9 (-1.7)           & 63.4 (-18.2)         & 68.9 (-12.7)        & 79.8 (-1.8)         & 81.6     \\                   
\bottomrule[0.35mm]
\end{tabular}}
\vspace{-15pt}
\end{center}
\caption{Ablation studies on the proposed IRLS scheme.}
\vspace{-5pt}
\label{tab:ablsync}
\end{table}


%% file: content/05_conclusion.tex
\section{Conclusion}
\label{sec:conclu}

In this paper, we propose a novel multiview point cloud registration method.
The key of the proposed method is a learning-based sparse pose graph construction which can estimate a overlap ratio between two scans, enabling us to select high-overlap scan pairs to construct a sparse but reliable graph. 
Then, we propose a novel history reweighting function in IRLS scheme, which improves robustness to outliers and has better convergence to correct poses.
The proposed method demonstrates the state-of-the-arts performances on both indoor and outdoor datasets with much fewer pairwise registrations.

%% file: content/06_supplementary.tex
\clearpage
\section{Appendix}

\setcounter{equation}{0}
\setcounter{table}{0}
\setcounter{figure}{0}
\setcounter{subsection}{0}
\renewcommand{\theequation}{A.\arabic{equation}}
\renewcommand{\thetable}{A.\arabic{table}}
\renewcommand{\thefigure}{A.\arabic{figure}}
\renewcommand{\thesubsection}{A.\arabic{subsection}}

In this supplementary material, we provide the detailed solution of pose synchronization in Sec.~\ref{sec:posesyn}, the implementation details in Sec.~\ref{sec:train}, additional analysis in Sec.~\ref{sec:more}, the running time analysis in Sec.~\ref{sec:time}, and more qualitative results in Sec.~\ref{sec:qual}.

\subsection{Pose synchronization}
\label{sec:posesyn}
In this section, we provide the detailed solution of pose synchronization in Sec.~3.3.2 of the main paper.
Given the edge weights and input relative poses $\{w_{ij},T_{ij}|(i,j) \in \mathcal{E}\}$, we solve the transformation synchronization by dividing it into rotation synchronization~\cite{arie2012global,huang2019learning} and translation synchronization~\cite{huang2017translation}. 
In the following, our pairwise transformation $T_{ij} = (R_{ij},t_{ij})$ on edge $(i,j) \in \mathcal{E}$ aligns the source scan $P_j$ to the target scan $P_i$. The scan poses are assumed to be camera-to-world matrices. Thus scans under the correctly recovered poses $\{(R_i,t_i)\}$ should reconstruct the whole scenario. 


\textbf{Rotation synchronization}.
Following~\cite{martinec2007robust,arie2012global,gojcic2020learning}, we treat the synchronization of rotations $\{R_i\}$ as an over-constrained optimization problem: 
\begin{equation}
    \mathop{\arg\min}_{R_1,...R_N \in SO(3)} \sum_{(i,j) \in \mathcal{E}} w_{ij}\|R_{ij} - R_i^TR_j\|^2_F,
    \label{eq:so3syn}
\end{equation}
where $\|\cdot\|_F$ means the Frobenius norm of the matrix.
Under the spectral relaxation, a closed-from solution of Eq.~\ref{eq:so3syn} can be computed as follows~\cite{arie2012global,gojcic2020learning}. Consider a symmetric matrix $L \in \mathbb{R}^{3N*3N}$ containing $N^2$ $3\times3$ blocks:
\begin{equation}\small
L = 
\left(              
  \begin{array}{cccc}  
    \sum\limits_{(1,j) \in \mathcal{E}}{w_{1j}} \mathbf{I}_3 & -w_{12}R_{12} & \cdots &  -w_{1N}R_{1N} \\ 
    -w_{21}R_{21} & \sum\limits_{(2,j) \in \mathcal{E}}{w_{2j}} \mathbf{I}_3 & \cdots &  -w_{2N}R_{2N} \\ 
    \vdots  & \vdots & \ddots &  \vdots \\ 
    -w_{N1}R_{N1} & -w_{N2}R_{N2} & \cdots   &  \sum\limits_{(N,j) \in \mathcal{E}}{w_{Nj}} \mathbf{I}_3\\
  \end{array}
\right),
\end{equation}
where $\mathbf{I}_3 \in \mathbb{R}^{3*3}$ denotes the identity matrix. For each edge $(i,j) \in \mathcal{E}$, we fill $-w_{ij}R_{ij}$ and $-w_{ij}R_{ij}^T$ to the $(i,j)$ and $(j,i)$ block. For unconnected edges, we set the corresponding blocks to zeros.

We first calculate three eigenvectors $\tau_1 , \tau_2, \tau_3 \in \mathbb{R}^{3N}$ corresponding to the three smallest eigenvalues $\lambda_1 < \lambda_2 < \lambda_3$ of $L$ and stack them to form $\gamma = [\tau_1,\tau_2,\tau_3] \in \mathbb{R}^{3N*3}$. Then, $v_i = \gamma[3i-3:3i] \in \mathbb{R}^{3*3}$ is an approximation of the absolute rotation $R_i$ for point cloud $P_i$ but may not satisfy the constraint $v_iv_i^T = \mathbf{I}_3$. Therefore, we rectify this by applying singular value decomposition on $v_i$ by $v_i = U_i\sum_iV_i^T$ and deriving $R_i = V_iU_i^T$~\cite{arie2012global}. Then, we further check $det(R_i)$ and exchange the first two rows of $R_i$ if $det(R_i)=-1$.

\textbf{Translation synchronization}.
Translation synchronization retrieves the translation vectors $\{t_i\}$ that minimize the problem:
\begin{equation}
    \mathop{\arg\min}_{t_1,...,t_N \in \mathbb{R}^3} \sum_{(i,j) \in \mathcal{E}} w_{ij}\|R_it_{ij}-t_j+t_i\|^2.
    \label{eq:r3syn}
\end{equation}
We solve it by the standard least square method~\cite{huang2017translation}.

Assuming $E$ edges are connected in $\mathcal{G}$, we thus construct three matrices $A$, $B$, and $H$ as follows. 
$A \in \mathbb{R}^{3E*3E}$ is initialized as an identity matrix.
$B \in \mathbb{R}^{3E*3N}$ contains $E*N$ $3\times3$ blocks and is initialized as a zero matrix. $H \in \mathbb{R}^{3E*1}$ is a vector containing $E$ $3\times1$ blocks. 
For the \textit{e-th} edge $(i,j) \in \mathcal{E}$, we multiply $A[3e-3:3e]$ with $w_{ij}$, fill $\mathbf{I}_3$ and -$\mathbf{I}_3$ to the $(e,j)$ and $(e,i)$ block of $B$ respectively, and fill $R_it_{ij}$ to the \textit{e-th} block of $H$. 
We thus solve $t = (B^TAB)^{-1}B^TAH$ and obtain the translation vector $t_i$ of each scan $P_i$ as $t[3i-3:3i]$.

\subsection{Implementation details}
\label{sec:train}

\begin{figure}
\begin{center}
\includegraphics[width=\linewidth]{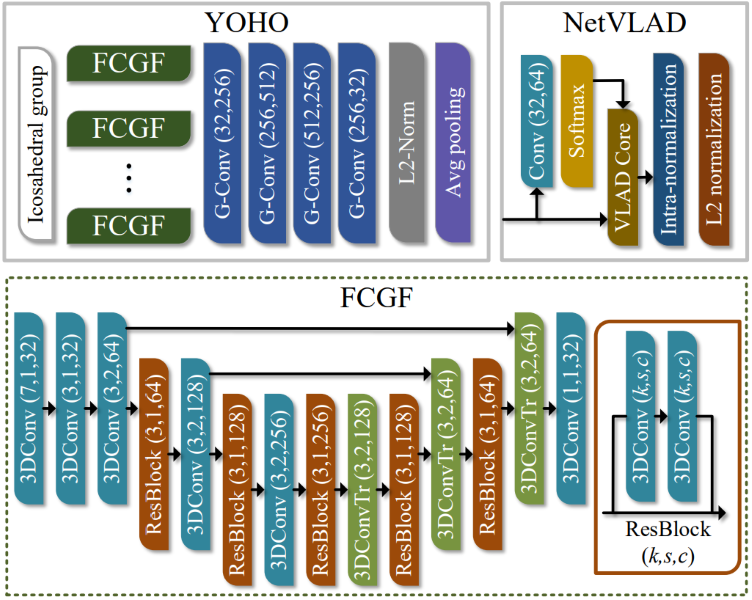}
\end{center}
\caption{Network architecture for global feature extraction. ``G-Conv'' means group convolution defined on Icosahedral group same as~\cite{wang2022you}. ``VLAD core'' is the same as~\cite{arandjelovic2018netvlad}. For FCGF~\cite{choy2019fully}, ``3DConv'' and ``3DConvTr'' denotes a sparse convolution layer and the transpose convolution layer for upsampling, respectively.}
\label{fig:arch}
\end{figure}

\subsubsection{Architecture}
The architecture of our global feature extraction network is shown in Fig.~\ref{fig:arch}. We adopt YOHO with the same architecture as~\cite{wang2022you} for 32-dim local feature extraction. More local feature extraction details can be found in~\cite{wang2022you}.
The extracted local features are aggregated to a global feature by a \textit{NetVLAD} layer~\cite{arandjelovic2018netvlad}. We set the number of clusters in \textit{NetVLAD} to 64 and the dimension of the global feature is thus 2048. Please refer to~\cite{arandjelovic2018netvlad} for more global feature aggregation details. 

\subsubsection{Training details}
We use the pretrained YOHO~\cite{wang2022you} for local feature extraction and train the $NetVLAD$ layer using the 46 scenes in the training split of 3DMatch~\cite{zeng20173dmatch}. We adopt the following data augmentations.
For each scene in the train set of 3DMatch, we first randomly sample $\alpha \in [8,60]$ scans as the graph node. Then, on each scan, we randomly sample $\beta \in [1024,5000]$ keypoints to extract YOHO features. The local features of $\alpha$ scans are fed to \textit{NetVLAD} to extract $\alpha$ scan global features. Then, we compute the $\binom{\alpha}{2}$ overlap scores by exhaustively correlating every two global features and compute the L1 distance between the ground-truth overlap ratios and the predicted overlap scores as the loss for training. 
We set the batch size to 1 and use the Adam optimizer with a learning rate of 1e-3. The learning rate is exponentially decayed by a factor of 0.7 every 50 epoch. In total, we train the $NetVLAD$ for 300 epochs.

\begin{table}
\begin{center}
\resizebox{0.7\linewidth}{!}{
\begin{tabular}{l|cc}
\toprule[0.35mm]
\textit{Overlap Estimation}  & \textit{3D-RR(\%)} & \textit{3DLo-RR(\%}) \\
\hline\hline
Predator~\cite{huang2021predator}  & 95.2    & 78.4      \\
Ours      & 96.2    & 81.6      \\  
\bottomrule[0.35mm]
\end{tabular}}
\end{center}
\caption{Registration recall on 3D(Lo)Match using estimated overlap scores from Predator~\cite{huang2021predator} and ours.}
\label{tab:predator}
\end{table}

\subsection{More analysis}

\label{sec:more}
\subsubsection{Use Predator~\cite{huang2021predator} for overlap estimation}
In Table.~\ref{tab:predator}, we use the overlap scores predicted by Predator~\cite{huang2021predator} in the sparse graph construction, which yields slightly worse results.
Moreover, Predator~\cite{huang2021predator} applies cross-attention layers between local features of a scan pair to estimate overlap while we only need to compute a global feature for every scan and efficiently correlate the global features to estimate overlap. In our test, the proposed method is 10$\times$ more efficient than Predator.

\subsubsection{Concurrent multiview registration works}
After our submission to CVPR 2023, two concurrent mulitview registration works are available online, namely, SynMatch~\cite{syncmatch} and HL-MRF~\cite{wu2023hierarchical}. SyncMatch and HL-MRF are specifically designed for registering raw RGB-D sequences and TLS point clouds, respectively, while the proposed method offering a more general approach. 
In our test, the proposed method notably outperforms SynMatch by $27\%$ on the 3DMatch dataset.
HL-MRF indeed performs well on the TLS-based ETH dataset but fails on the indoor datasets.

\subsubsection{Estimated overlap versus ground truth overlap}

\begin{figure}
\begin{center}
\includegraphics[width=0.7\linewidth]{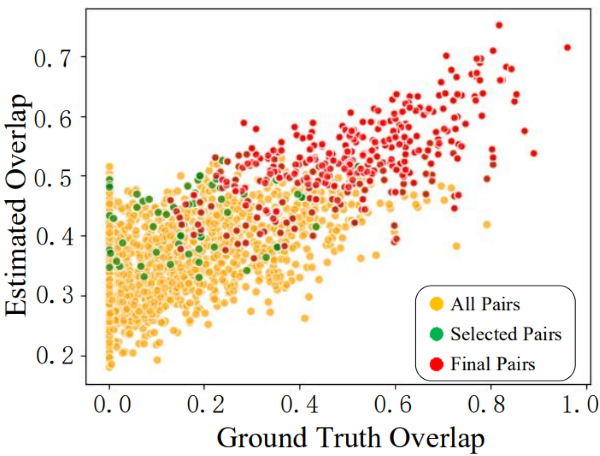}
\end{center}
\caption{Estimated overlap ratio versus the ground truth overlap ratio on scan pairs of the \textit{Kitchen} scene of 3DMatch. ``All pairs'' means all $\binom{N}{2}$ scan pairs. ``Selected pairs'' means the scan pairs selected to construct the sparse pose graph. ``Final pairs'' means the scan pairs with an edge weight greater than $10^{-2}$ after transformation synchronization.}
\label{fig:gt_pre}
\end{figure}

In Fig.~\ref{fig:gt_pre}, each point $(o_{gt},o_{est})$ represents a scan pair with the ground truth overlap ratio $o_{gt}$ and estimated overlap ratio $o_{est}$. The plot reveals several observations: (1) scan pairs with larger ground truth overlaps indeed have larger overlap scores; (2) the constructed sparse graph mainly contains scan pairs with higher overlap ratios, as evidenced by the green and red points; (3) the proposed transformation synchronization algorithm further eliminates unreliable scan pairs effectively to achieve accurate scan poses, as shown by the red points.

\subsubsection{Use different \textit{top-k} in sparse graph construction}
In Table.~\ref{tab:topfull}, we show the results with different $k$ in the sparse graph construction.
Retaining too many scan pairs with larger $k$ may include more outliers while using too small $k$ could split the whole graph into several disconnected subgraphs. Results show that using $k=10$ or 12 brings the best results.

\begin{table}
\begin{center}
\resizebox{\linewidth}{!}{
\begin{tabular}{l|cccccc|c}
\toprule[0.35mm]
\textit{Top-}           & 4    & 6    & 8    & 10   & 12   & 15   & Full  \\\hline \hline
\textit{\# Pair }   & \textbf{1167} & 1707 & 2250 & 2798 & 3349 & 4129 & 11905 \\
\textit{Sync-time} (s)   & \textbf{20.2}   & 30.4   & 37.4   & 54.8   & 66.6   & 90.3   & 405.4   \\
\textit{3D-RR} (\%)    & 91.3 & 91.6 & 95.5 & 96.2 & \textbf{96.6} & 96.0 & 93.2  \\
\textit{3DL-RR} (\%)  & 71.0 & 74.7 & 80.9 & \textbf{81.6} & 81.2 & 80.3 & 76.8  \\
\bottomrule[0.35mm]
\end{tabular}}
\end{center}
\caption{Ablation study on \textit{k} in sparse graph construction. ``Full'' means using fully-connected graphs. ``Sync-time'' means the time for transformation synchronization.}
\label{tab:topfull}
\end{table}

\subsubsection{Performances using different IRLS iterations}
\begin{figure}
\begin{center}
\includegraphics[width=0.8\linewidth]{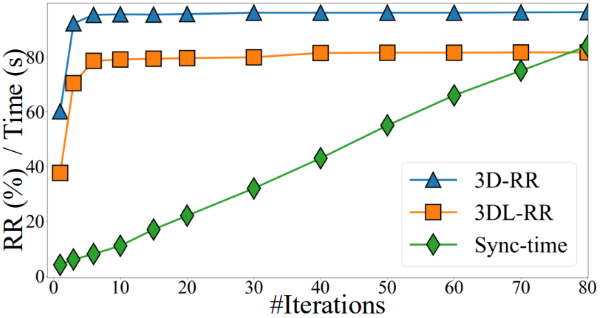}
\end{center}
\caption{Results of the proposed history reweighting IRLS with different iterations.}
\label{fig:iters}
\end{figure}

In Fig.~\ref{fig:iters}, we show the registration performance on 3D(Lo)Match with different iteration numbers. It can be seen that the results will be better with more iterations. However, using more iterations also costs more time. We thus select 50 iterations for its stable performance and efficiency by default.

\begin{table}
\begin{center}
\resizebox{0.9\linewidth}{!}{
\begin{tabular}{l|cc|c}
\toprule[0.35mm]
\multicolumn{1}{c|}{\textit{Method}} & \textit{Graph Cons} (s) & \textit{Trans Sync} (s) & \textit{Total} (s) \\\hline\hline
RotAvg~\cite{chatterjee2017robust} + Full             & 86.3  & 49.4 & 135.7 \\
LITS~\cite{yew2021learning} + Full                & 86.3  & 0.7  & 87.0  \\
HARA~\cite{lee2022hara} + Full                & 87.3  & 8.5  & 95.8  \\
\cdashline{1-4}[3pt/3pt]
RotAvg~\cite{chatterjee2017robust} + Pruned~\cite{gojcic2020learning}         & 164.1 & 22.6 & 186.7 \\
LITS~\cite{yew2021learning} + Pruned~\cite{gojcic2020learning}            &  164.1 & \textbf{0.7}  & 164.8 \\
HARA~\cite{lee2022hara} + Pruned~\cite{gojcic2020learning}              & 164.8 & 7.5  & 172.4 \\
\cdashline{1-4}[3pt/3pt]
Ours                       &\textbf{20.0}  & 6.9  & \textbf{26.8}     \\
\bottomrule[0.35mm]
\end{tabular}}
\end{center}
\caption{Detailed time consumption for registering a scene on 3DMatch. ``Graph Cons'' means the time for constructing the input pose graph. ``Trans Sync'' means the time for IRLS-based transformation synchronization.}
\label{tab:time}
\end{table}

\subsection{Runtime analysis}
\label{sec:time}
In Table.~\ref{tab:time}, we provide the runtime for the graph construction and the IRLS-based transformation synchronization averaged on the 8 scenes of the 3DMatch dataset.
We evaluate the runtimes on a computer with Intel(R) Core(TM) i7-10700 CPU@ 2.90GHz with GeForce GTX 2080Ti and 64 GB RAM.
Our sparse pose graph construction is nearly $67s$ faster than baselines for conducting much fewer pairwise registrations. In total, our method is $61s \sim 160s$ faster than baselines for registering a scene in 3DMatch.

\subsection{Limitations}
\begin{figure}
\begin{center}
\includegraphics[width=0.8\linewidth]{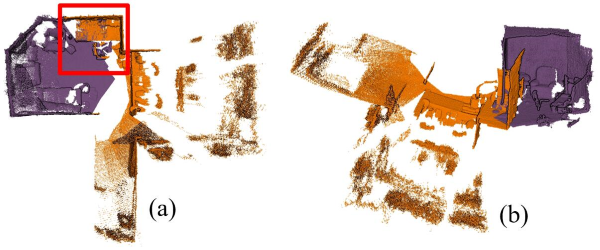}
\end{center}
\caption{A failure case in ScanNet. (a) The ground truth multiview registration (30 scans). (b) The multiview registration from the proposed method.}
\label{fig:fail}
\end{figure}
When the overlap ratios of two scans are too small and there are no other scans which forms a cycle with these two scans, our method may fail in this case. A typical example is shown in Fig.~\ref{fig:fail}, where overlap region in the red rectangle is very small and mainly consists of
feature-less planar points. In this case, our method fails to register the whole scene but separately recover poses on two subgraphs. This also shows that our method may have the potential to automatically separate scans from two different scenes, which is beyond the discussion of this paper.

\subsection{More qualitative results}
\label{sec:qual}
\begin{figure*}
\begin{center}
\includegraphics[width=0.9\linewidth]{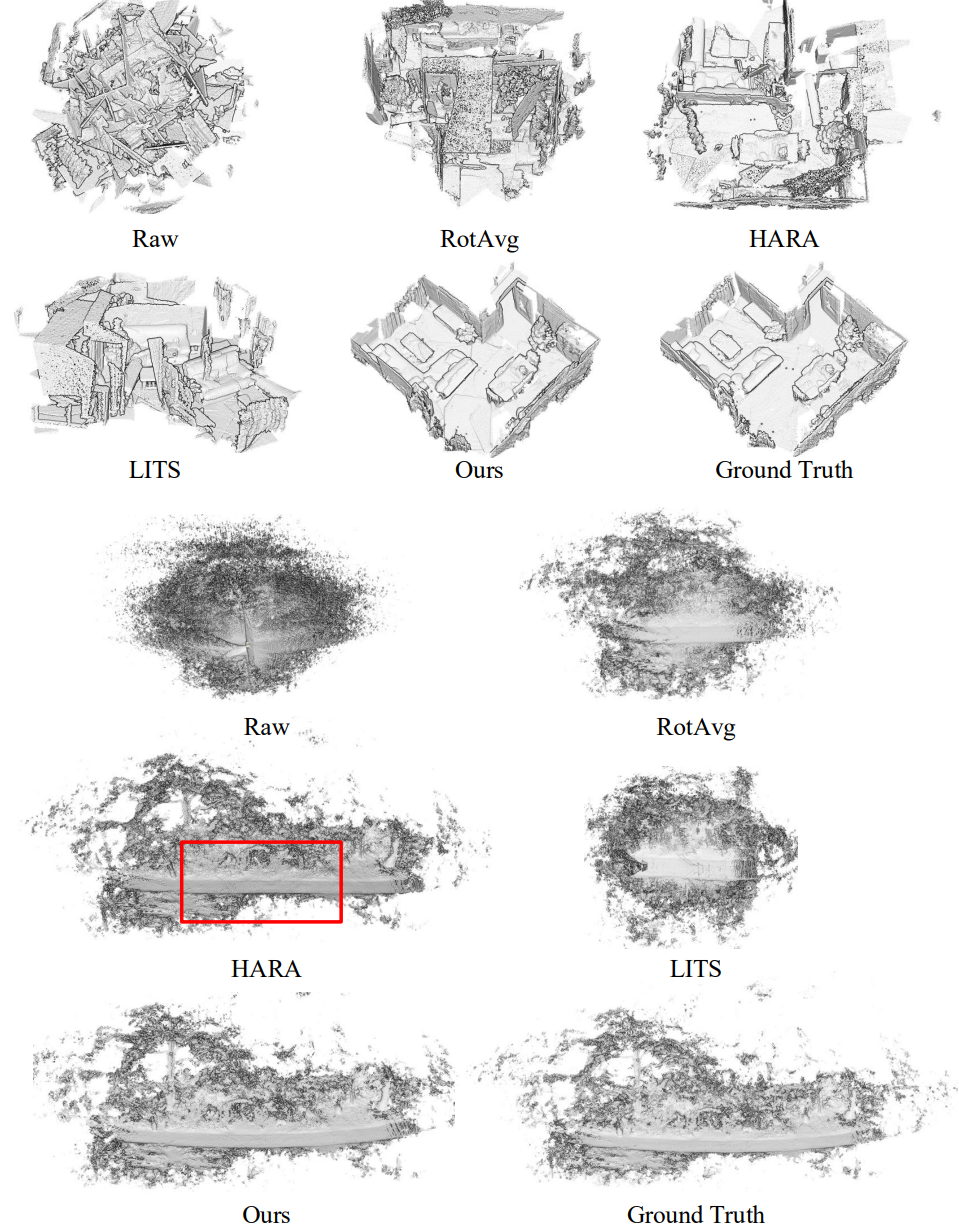}
\end{center}
\caption{Registration results of our method, RotAvg~\cite{chatterjee2017robust}, HARA~\cite{lee2022hara}, and LITS~\cite{yew2021learning} on the 3DMatch dataset and the ETH dataset. Top: the \textit{Home1} scene of 3DMatch. Bottom: the \textit{Wood\_Summer} scene of ETH.}
\label{fig:3dm}
\end{figure*}

\begin{figure*}
\begin{center}
\includegraphics[width=0.9\linewidth]{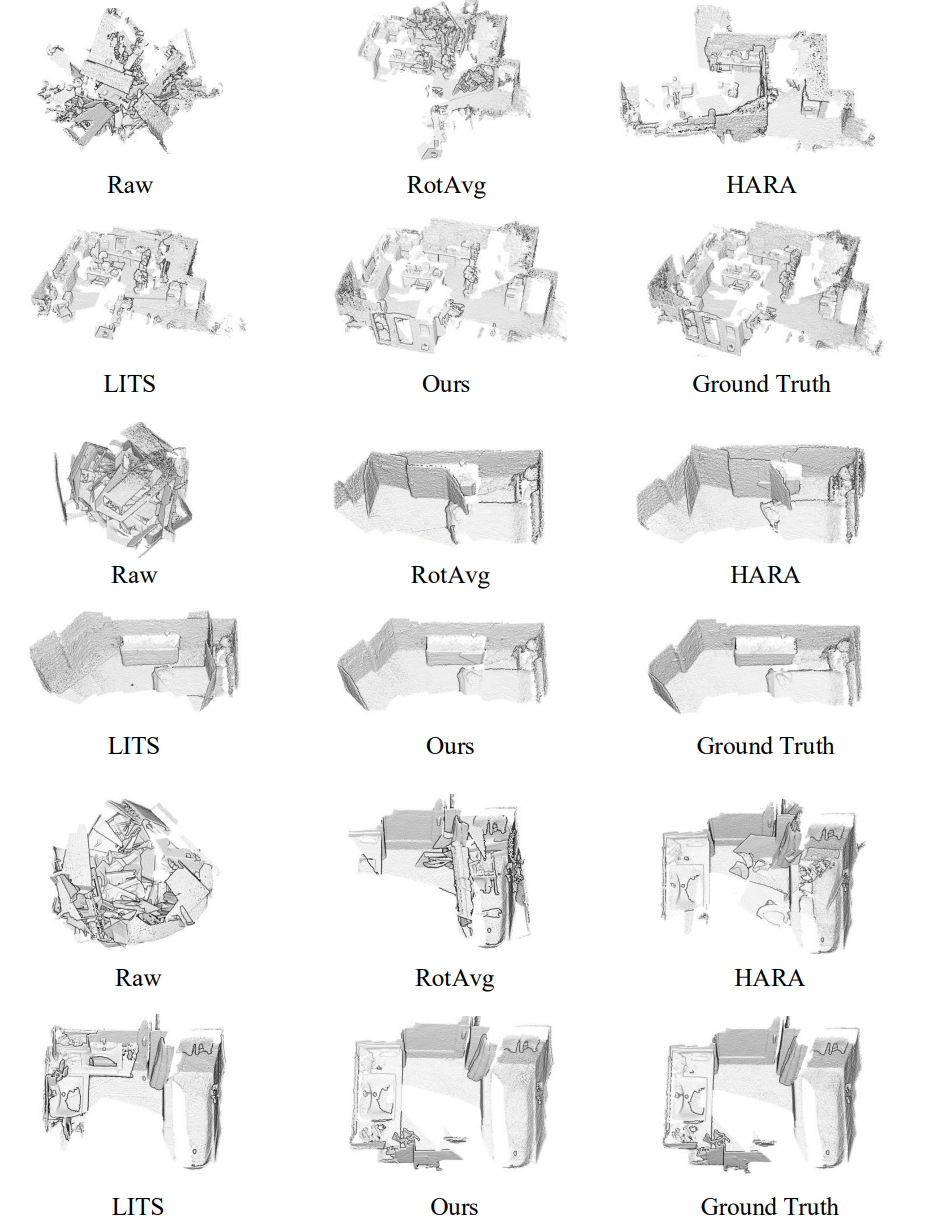}
\end{center}
\caption{Registration results of our method, RotAvg~\cite{chatterjee2017robust}, HARA~\cite{lee2022hara}, and LITS~\cite{yew2021learning} on scenes of ScanNet dataset including \textit{Scene0309\_00} (top), \textit{Scene0286\_02} (middle), and \textit{Scene0265\_02} (bottom).}
\label{fig:scannet}
\end{figure*}

\begin{figure*}
\begin{center}
\includegraphics[width=0.9\linewidth]{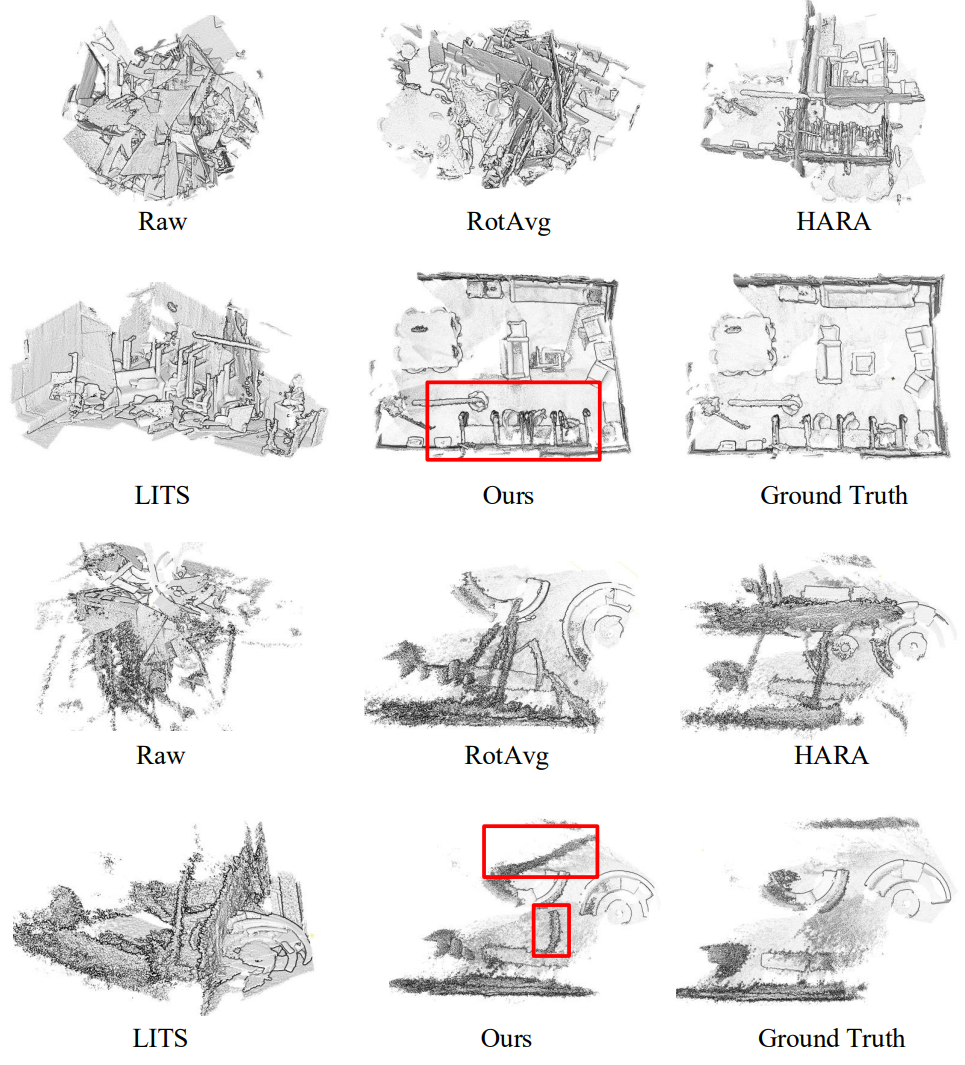}
\end{center}
\caption{Registration results of our method, RotAvg~\cite{chatterjee2017robust}, HARA~\cite{lee2022hara}, and LITS~\cite{yew2021learning}  on 3DMatch (top: \textit{Studyroom}) and ScanNet (bottom: \textit{Scene0334\_02}). Our method fails to register the scans in the red boxes due to repetitive structures and feature-less regions.}
\label{fig:failsupp}
\end{figure*}

We provide additional qualitative results including success cases (Fig.~\ref{fig:3dm} and Fig.~\ref{fig:scannet}) and failure cases (Fig.~\ref{fig:failsupp}). We also compare our results with the registration results of RotAvg~\cite{chatterjee2017robust}, HARA~\cite{lee2022hara}, and LITS~\cite{yew2021learning}. The failure of our method occurs when some overlap regions mainly contain the repetitive structures (top of Fig.~\ref{fig:failsupp}) or feature-less regions (bottom of Fig.~\ref{fig:failsupp}).

